\documentclass[a4paper, 10pt, conference]{ieeeconf}      

\IEEEoverridecommandlockouts                              

\usepackage{times} 
\usepackage{amsmath} 
\usepackage{amssymb}  

\usepackage{graphicx}
\graphicspath{ {images/} {data/} }

\usepackage{verbatim}
\usepackage{siunitx}
\usepackage{multirow}
\usepackage{tabularx}
\usepackage{makecell}
\usepackage{caption}
\usepackage{subcaption}

\usepackage{xspace}

\makeatletter
\DeclareRobustCommand\onedot{\futurelet\@let@token\@onedot}
\def\@onedot{\ifx\@let@token.\else.\null\fi\xspace}
\def\eg{\emph{e.g}\onedot}

\def\ie{\emph{i.e}\onedot}

\newsavebox\myboxA
\newsavebox\myboxB
\newlength\mylenA
\newcommand*\xoverline[2][0.75]{%
	\sbox{\myboxA}{$\m@th#2$}%
	\setbox\myboxB\null
	\ht\myboxB=\ht\myboxA%
	\dp\myboxB=\dp\myboxA%
	\wd\myboxB=#1\wd\myboxA
	\sbox\myboxB{$\m@th\overline{\copy\myboxB}$}
	\setlength\mylenA{\the\wd\myboxA}
	\addtolength\mylenA{-\the\wd\myboxB}%
	\ifdim\wd\myboxB<\wd\myboxA%
	\rlap{\hskip 0.5\mylenA\usebox\myboxB}{\usebox\myboxA}%
	\else
	\hskip -0.5\mylenA\rlap{\usebox\myboxA}{\hskip 0.5\mylenA\usebox\myboxB}%
	\fi}

\makeatother

\newcommand*\rot{\rotatebox{90}}

\title{\LARGE \bf
Data Selection for training Semantic Segmentation\\ CNNs with cross-dataset weak supervision
}

\author{Panagiotis Meletis$^{}$, Rob Romijnders, and Gijs Dubbelman$^{}$
	\thanks{$^{}$Panagiotis Meletis ({\tt\small p.c.meletis@tue.nl}), Gijs Dubbelman ({\tt\small g.dubbelman@tue.nl}), and Rob Romijnders ({\tt\small romijndersrob@gmail.com}) are with the Department of Electrical Engineering, Eindhoven University of Technology,
	Eindhoven, The Netherlands.
	This research has received funding from ECSEL JU in collaboration with the European Union's H2020 Framework Programme and National Authorities, under grant agreement no. 783190.
    }%
}

\begin{document}

\maketitle
\thispagestyle{empty}
\pagestyle{empty}

\begin{abstract}
Training convolutional networks for semantic segmentation with strong (per-pixel) and weak (per-bounding-box) supervision requires a large amount of weakly labeled data. We propose two methods for selecting the most relevant data with weak supervision. The first method is designed for finding \textit{visually similar} images without the need of labels and is based on modeling image representations with a Gaussian Mixture Model (GMM). As a byproduct of GMM modeling, we present useful insights on characterizing the data generating distribution. The second method aims at finding images with high \textit{object diversity} and requires only the bounding box labels. Both methods are developed in the context of automated driving and experimentation is conducted on Cityscapes and Open Images datasets. We demonstrate performance gains by reducing the amount of employed weakly labeled images up to 100 times for Open Images and up to 20 times for Cityscapes.
\end{abstract}

\section{Introduction and Related Work}
Recently, multiple dataset training of convolutional networks is gaining attention \cite{fourure2016semantic, meletis2018heterogeneous, rob2018rob}, since it offers improved performance and better generalization capabilities compared to single dataset training. Multiple dataset training is especially advantageous for training semantic segmentation networks, which requires large amounts of training examples~\cite{long2015fully}. However, factors as different dataset sizes, repetitive examples (low informative value), and high annotation costs, hamper the effectiveness of multiple dataset training. These factors especially influence methods that employ weaker forms of supervision \cite{ye2018learning, xu2015learning, kumar2011learning, meletis2018heterogeneous}.

The current trend to deal with the aforementioned challenges is model selection,~\ie design, train and tune a convolutional network for robustness and performance. A less studied research branch, data selection~\cite{birodkar2019semantic, vodrahalli2018all, liu2018pixel} appears more appealing. In this work, we propose two data selection methods, which indicate how data should be chosen for maximizing \textit{visual similarity} and \textit{object diversity} among used datasets, and are well suited for multiple dataset training. The first method, employs a Gaussian Mixture Model (GMM) in order to model image representations of a dataset, and the second one uses predefined scoring heuristics to rank images.

Our data selection methods can be employed in cases where: 1) fewer data need to be used, by selecting the most informative images, 2) fewer data need to be annotated, by selecting most similar images between labeled and unlabeled datasets, and 3) a balanced amount of examples between datasets is preferred (\eg for multiple dataset training).

In this work we focus in the problem of training a semantic segmentation model using strong (per-pixel) and weak (per-bounding-box) supervision from different datasets and we show the benefits of the proposed data selection schemes both in performance and in decreasing the number of needed examples. Specifically, data selection allows the a model to reach same levels of performance using 10 to 100 times less weakly annotated data. Furthermore, we present results towards modeling the visual domain of a dataset and quantifying \textit{visual similarity} and \textit{object diversity}.

\begin{figure}
	\begin{center}
		\includegraphics[width=1.0\linewidth]{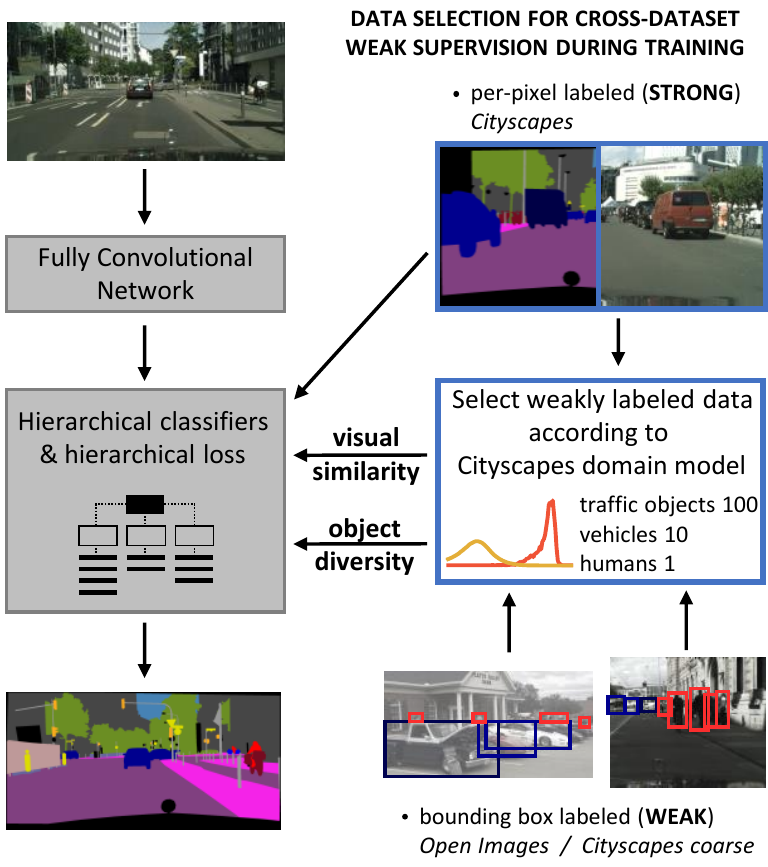}
	\end{center}
	\caption{The proposed selection methods aim at selecting image-label pairs from the weakly labeled datasets that are visually similar to the strongly labeled dataset and contain high diversity of objects of interest. Subsequently, by training on both strongly and selected weakly labeled datasets we show the benefits of data selection for semantic segmentation.}
	\label{fig:eye-catcher}
\end{figure}

To summarize, in this work we:
\begin{itemize}
	\item propose a selection method, based on modeling image representations with a GMM, for finding \textit{visually similar} images to a given dataset,
	\item propose a selection method, based on class scoring heuristics, for finding rich labeled images,
	\item apply both methods independently and jointly in weak supervision selection for semantic segmentation to reduce the amount of required training examples while increasing performance, and
	\item present results towards characterizing the image domain of a dataset through GMM modeling.
\end{itemize}

Our trained convolutional networks, the GMM models, and the two selection methods algorithms will be made available to the research community~\cite{panos2019code}.

\begin{figure*}
	\centering
	\includegraphics[width=1.0\linewidth, trim={0.0cm 27.5cm 0.0cm 0.0cm}, clip]{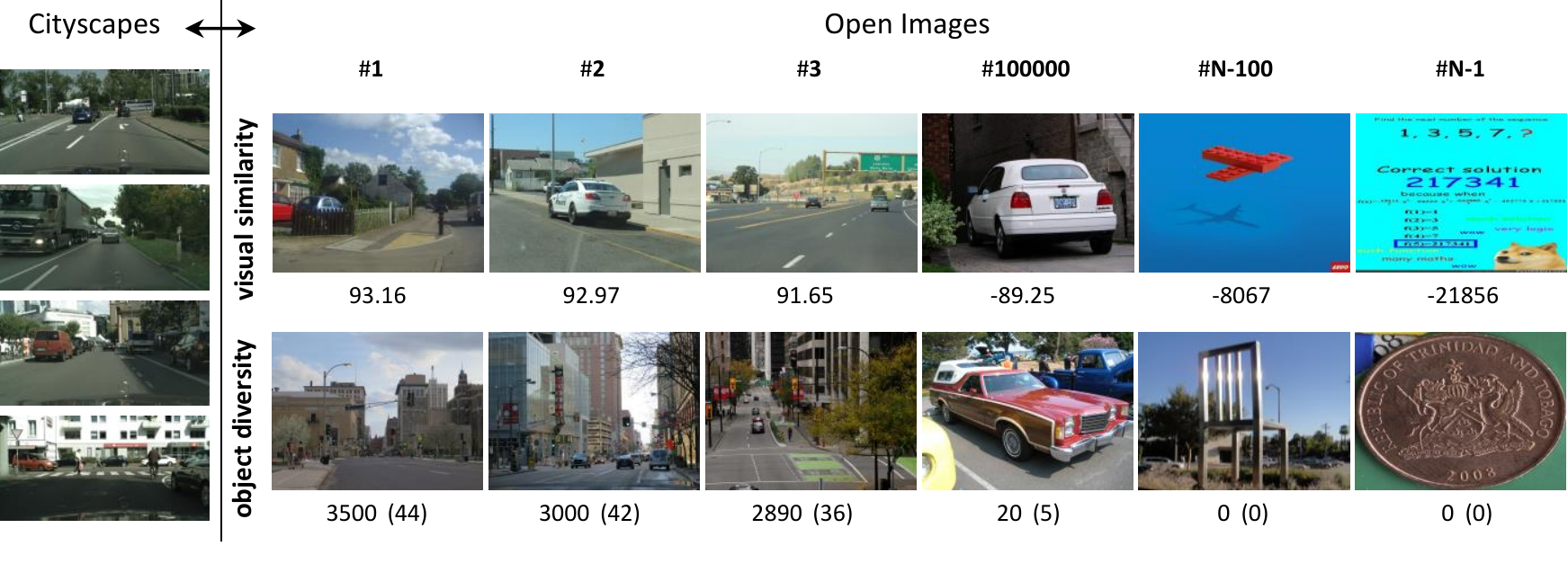}
	\caption{Example of selected images from $N = 1.74$ million Open Images images using our data selection methods in descending order. First row: \textit{visual similarity} using GMM, the $sim_{citys}$ measure is shown. Second row: \textit{object diversity} using class scores, the heuristics scores and the number of objects of interest are shown.}
	\label{fig:sel-ims}
\end{figure*}

\section{Problem Definition}
\label{sec:prob-def}
Semantic Segmentation is a pixel-level task and as such, it requires a large amount of per-pixel labeled images that are hard to obtain. This is specifically costly in the automated driving field, where small sized per-pixel labeled datasets are available. Although much larger datasets exist for complementary tasks to semantic segmentation, like object detection or classification, they are not specialized on street scenes, but contain generic complex scenes. Thus, apart from the different type of annotations, we have to deal also with the \textit{domain gap} between the chosen datasets, since it is also prefered the trained model to have good generalization properties. The methods that have been proposed in the literature attempt to take advantage of these datasets with a weakly or semi-supervised learning approach.

In this paper, we assume that we have
\begin{enumerate}
    \item a labeled dataset $C = \{(x, y)_i, ~ i = 1, ..., N\}$ with $N$ pairs of images $x$ and per-pixel labels $y$ for semantic segmentation,
    \item a dataset $O = \{(x[, z])_i, ~ i = 1, ..., M\}$ with $M$ images $x$ and optionally $M$ weak labels $z$ from the perspective of semantic segmentation (\eg bounding boxes or image-level labels), where $M \gg N$, and
    \item a convolutional network model that can be trained on strong and weakly labeled datasets,
\end{enumerate}
and we seek a methodology for selecting images from $O$ that are \textit{visually similar} to $C$ and informative enough for $y$.

The data selection problem requires selecting images from $O$ that are as \textit{visually similar} to images in $C$ as possible. This is desired, since the \textit{domain gap}, which can be seen as the dissimilarity in image content and appearance, may hinder the training procedure in the multiple dataset setting. In the same time, it is also preferred that the images have high informative value for the classes that we want to train for, or in other words to have high \textit{object diversity}.

In this work we experiment with Cityscapes Dense subset~\cite{Cordts2016Cityscapes} as $C$, and Cityscapes Coarse subset and Open Images bounding boxes subset~\cite{kuznetsova2018open} as $O$. We employ the hierarchical convolutional networks of ~\cite{meletis2019boosting} that can be trained on multiple datasets with strong and weak labels, which require the labels $z$ of dataset $O$.

\section{Method}
\label{sec:method}
In this Section we describe our two proposed data selection methods, how they can be combined, and their connection with the notions of \textit{visual similarity} and \textit{object diversity}. In this work, we aim at discovering \textit{visual similarity} using only the images and not the associated labels, and \textit{object diversity} using only the labels and not the images. We experiment on three datasets, namely Open Images, Cityscapes Dense, and Cityscapes Coarse (see Section~\ref{ssec:datasets} for more details).

The goal of our methods is to select images from the weakly labeled datasets (Open Images, Cityscapes Coarse), that are \textit{visually similar} and have high \textit{object diversity} compared to the strongly labeled dataset (Cityscapes Dense).

\subsection{Gaussian Mixture Model: visual similarity}
\label{ssec:method-gmm}
Inspired by ~\cite{birodkar2019semantic, robotka2009image} our method consists of three distinct phases that are described in the following Sections. First, we use a pre-trained convolutional network to extract a low dimensional representation for each Cityscapes Dense image, then we fit a GMM to those representations, and finally we use that model to rank the images of the weakly labeled datasets. We hypothesize that images that are \textit{visually similar} to Cityscapes Dense,~\ie depict street scenes, will have high probability density under the GMM and images from generic scenes,~\ie the majority of Open Images images not containing street scenes, will have low probability density.

\subsection*{Extracting image representations}
\label{ssec:feat-extract}
We aim to capture the distribution of Cityscapes Dense image domain. Unfortunately, it is hard to fit probabilistic models to images in general \cite{kingma2013auto}, \cite{dumoulin2016adversarially}, and current state of the art in generative modelling of images does not assign calibrated density \cite{goodfellow2014generative}. As such, we extract representations from a fully convolutional network trained for semantic segmentation on Cityscapes Dense. The first layers of the trained neural network will serve for the extraction of representations, as we know that initial layers of a neural network maintain information about the input images \cite{tishby2000information}.

Neural networks are known to learn internal representations irrespective of the task they are trained on \cite{zamir2018taskonomy}, \cite{bengio2013representation}. Thus, training could involve any computer vision task, such as classification, segmentation, or detection, but we leave this for investigation in future research. In this work, we use a convolutional network to extract the image representations and we choose to train it for semantic segmentation, as this is the task where the selected images will be eventually used.

The backbone consists of a ResNet-50~\cite{he2016deep}, which we modify for semantic segmentation as in~\cite{meletis2018heterogeneous}. We choose to extract features from the penultimate convolutional layer, which has shape $\left( H, W, C \right)$, and we call this subnetwork $f$. If $x_i$ is the input image, the convolutional representation can be denoted as the set
\begin{equation}
\label{eq:repr-set}
\Phi_i = \left\lbrace f_{h, w, :} \left( x_i \right), ~ \forall ~ h \in H, w \in W \right\rbrace
\end{equation}
where $h, w$ index all the receptive fields of the penultimate layer corresponding to different regions on the input image $x_i$. In other words we slice the output of $f$, to the set $\Phi$ containing $H \cdot W$ elements with $C$ features (depth) each.

\subsection*{Modeling image representations}
We want to fit a probabilistic model to the low dimensional representations $\Phi_i$ for all images $x_i$ of dataset $C$ (see Section~\ref{sec:prob-def}). Such model would assign large probability density to the representations of the modeled domain (Cityscapes Dense), and low density to images outside this domain.

Here we choose a Gaussian Mixture Model (GMM)~\cite{mclachlan2019finite} for its simplicity and explicitness. Since assigning probability densities to entire Cityscapes images would be too costly task, we assume that for every image the set $\Phi_i$ contains independent and identically distributed representations and we average its elements:
\begin{equation}
\label{eq:avg-repr-set}
\xoverline{\Phi}_i = \frac{1}{|\Phi_i|} \sum_{e \in \Phi_i} e
\end{equation}
The next step is to model with a GMM the average representations $\xoverline{\Phi}_i$ for all images. A GMM is a mixture of K Gaussian distributions, with variable mixture coefficients $\pi_j$, means $\mu_j$, and variances $\sigma_j^2$ for the Gaussian distributions. We group those parameters into $\Psi = \left\lbrace \pi_1, ..., \pi_K, \mu_1, ..., \mu_K, \sigma_1, ..., \sigma_K \right\rbrace $. The log likelihood function for $\Psi$, given the independent average representations $\xoverline{\Phi}_i$ for all images $N$ of Cityscapes Dense, can be expressed as
\begin{equation}
\label{eq:log-likelihood}
\log L\left(\Psi\right) = \sum_{i=1}^{N} \log \sum_{j=1}^K \pi_j \mathcal{N}(\xoverline{\Phi}_i ~ ; ~ \mu_j, \sigma_j^2)
\end{equation}
The Maximum Likelihood estimate $\Psi_{citys}$ is found using Eq.~\ref{eq:log-likelihood} and the Expectation Maximization algorithm~\cite{mclachlan2019finite}.

\subsection*{Image to dataset visual similarity}
We define a measure of similarity to the domain that is modeled by the GMM,~\ie Cityscapes Dense, so we can rank images from others datasets, as the maximum log probability under the model for all receptive fields of an image $x_i$:
\begin{equation}
\label{eq:sim-measure}
sim_{citys}(x_i) = \max ~ \log p\left( \Phi_i ~ ; ~ \Psi_{citys} \right)
\end{equation}
and according to our hypothesis the larger $ sim $ is, the image is visually more similar to the modeled Cityscapes Dense image domain.
We rank weakly labeled images using $sim_{citys}$ in descending order and we select various top portions for the experiments of Sections~\ref{ssec:perf-open},~\ref{ssec:perf-citys}.

\subsection{Class score heuristics: object diversity}
A training image has high \textit{\textit{object diversity}} when it contains a large variety and number of objects of interest. In the context of automated driving we define three categories of objects, namely traffic objects (traffic signs and traffic lights), vehicles (car, truck, bus, motorcycle, bicycle, train), and humans (pedestrian, rider) and we assign to each category a score. These scores are defined by empirical tests and manual inspection of the images, and they depend on each dataset. The general intuition behind scoring is that traffic objects are most probable to appear in street scenes only, while vehicles and humans can appear in a variety of other scenes.

For Open Images, all above categories are labeled in a instance-wise manner and we assign 100, 10 and 1 points to them respectively. For Cityscapes traffic objects are not labeled instance-wise, so we assign weights for the last two categories, as 10 and 1 respectively. For each image the total score from all labeled objects is accumulated. The images are ranked according to their score and different top portions are selected for the experiments of Sections~\ref{ssec:perf-open},~\ref{ssec:perf-citys}.  

\subsection{Combine the two selection methods}
In the previous two Sections we described the two selection schemes and how they result in two rankings of the images of a dataset. In general the two rankings can have a different ordering, thus aggregating them into one collection is not a trivial task. Since, we equally prefer \textit{visual similarity} and \textit{object diversity} we opt for interleaving the rankings by interchangeably choosing images from the initial rankings to the final selection. In the process, if an image is already inserted in the final selection it is not inserted twice.


\section{Implementation details}
\label{sec:details}
In this Section, we describe the chosen convolutional model for training simultaneously on datasets with strong and weak supervision, we present the used datasets, and we provide the hyperparameters employed in training to enable reproducibility of our experiments.

\subsection{Datasets}
\label{ssec:datasets}
\textbf{Cityscapes Dense}:
Cityscapes dataset \cite{Cordts2016Cityscapes} contains street scene images from German cities taken from a 2 Mpixel camera mounted on a car. We used the training subset with 2975 densely (per-pixel) labeled images and the bigger subset of 20000 coarsely (per-pixel) labeled images.

\textbf{Open Images v4}: This dataset \cite{kuznetsova2018open} contains 9 million images from everyday, complex scenes collected from the internet and has multiple resolutions, shooting angles, and several objects that are not relevant for automated driving. The official subset labeled with 14.6 million bounding boxes contains 1.74 million images.

\textbf{Cityscapes Coarse bboxes}:
This a dataset with bounding boxes that was created for the purpose of this paper from the coarse, per-pixel, instance labels of Cityscapes Coarse subset. Specifically, for each labeled object in an image we define a bounding box using the minimum and maximum coordinates of per-pixel labels in each axis.

\subsection{Convolutional model for training on strong and weak supervision}
We use our published hierarchical convolutional network for training simultaneously on weak and strong supervision for semantic segmentation~\cite{meletis2018heterogeneous, meletis2019boosting}. The network consists of a conventional ResNet-50 feature extractor, that is modified to have semantic segmentation output with dilated convolutions and an upsampling module. Moreover, instead of one per-pixel classifier it consists of a hierarchy of classifiers arranged in a tree structure according to the classes hierarchy. Each of the classifiers is fed with the same convolutional features of the feature extractor. During inference, the results from all classifiers are aggregated in a per-pixel manner to output the final decisions.

\subsection{Training details}
For fair comparisons we train all networks in Section~\ref{sec:experiments} with the same hyperparameters and for the same number of epochs. For the image representation extraction of Section~\ref{ssec:method-gmm} we use input image dimensions of 1024 by 2048, which are reduce to a grid of 256 by 512 receptive fields, each observing an area of approximately 200 by 200 pixels on the full image. The representation's depth is 256.

In training the GMM, we sample from the $256 \cot 512 \cdot 2975 \approx 390 \cdot 10^6$ 256-dimensional representations only 24k from all the images in the Cityscapes Dense training set. For the GMM model, we fit the parameters of the mixtures using Expectation Maximization. We continue updates until the likelihood does not change from one E step to another by more than 0.001 nat. We use the open source implementation of the Scikit learn library \cite{scikit-learn}.

\section{Experiments}
\label{sec:experiments}
First, we present the overall results in Section~\ref{ssec:perf-over} for our two selection methods applied on two diverse datasets and we analyze them in Sections~\ref{ssec:perf-citys},~\ref{ssec:perf-open}. Then in Section~\ref{ssec:analysis}, we perform ablation experiments for the parameters of the models and we present an analysis and intuitions behind our methods. All IoU results refer to training a hierarchical segmentation network, as described in Section~\ref{sec:details}, on Cityscapes Dense (per-pixel labels) and on the weakly labeled (per-bounding-box) dataset (Cityscapes Coarse or Open Images). We evaluate the proposed data selection methods on the per-pixel labeled Cityscapes validation set unless otherwise noted. We use the Intersection over Union metric~\cite{long2015fully}. The IoU results are averaged over the last 5 (Cityscapes) epochs, when the model converges, since the variance is high. The mIoU results are the mean IoU over the classes that receive extra supervision from the weakly labeled dataset.

\begin{figure}
	\centering
	\includegraphics[width=1.0\linewidth]{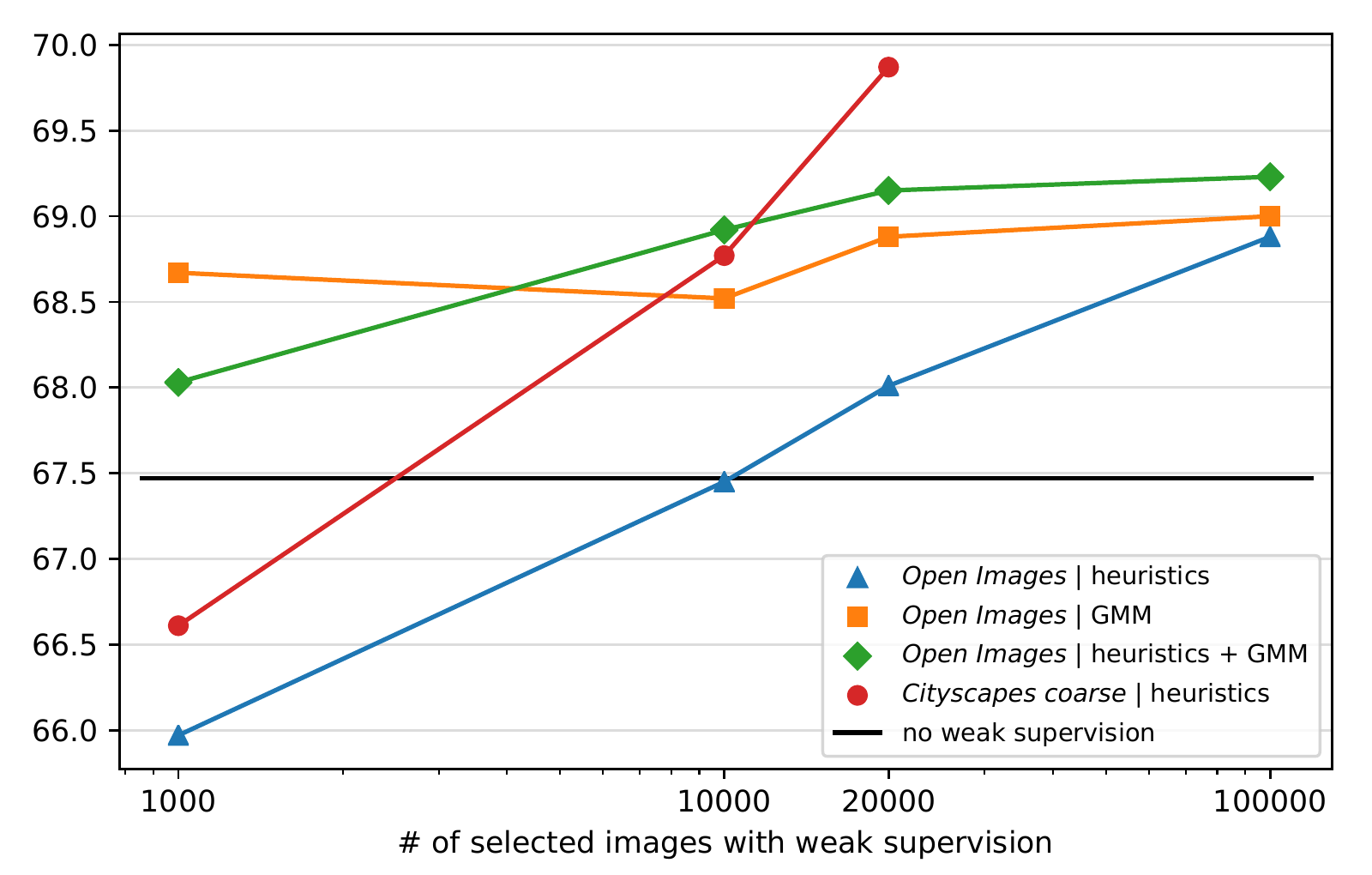}
	\caption{Performance (\textbf{mIoU}) on Cityscapes validation set. The networks are trained on Cityscapes Dense and optionally on additional selected data from Cityscapes Coarse and Open Images. The dots mark the conducted experiments. The black horizontal line denotes the mIoU of training without weak supervision.}
	\label{fig:perf-overall}
\end{figure}

\subsection{Overall results}
\label{ssec:perf-over}
In Figure~\ref{fig:perf-overall} the mIoU performance on Cityscapes validation set for various combinations of our selection methods and datasets is shown. We experiment on two different datasets. Cityscapes Coarse is a subset of Cityscapes and as such contains images of street scenes all shot from a specific point of view. Open Images is a generic scene dataset, collected from various image sources and point of views, and street scenes are rare. We demonstrate that data selection from both datasets is beneficial, by improving performance, while reducing the amount of required data.

From Figure~\ref{fig:perf-overall} we observe that class scoring heuristics, is advantageous for both datasets. As anticipated, the mIoU for the same amount of selected data, in the case of Cityscapes Coarse is always higher than Open Images, since the former dataset has \textit{visually similar} images to Cityscapes validation set than the latter. Interestingly, when the selection quantity is limited to 1000 images, scoring heuristics perform even lower than the baseline. This is unintuitive since we add data over the baseline, but it can be explained because we use exactly the same hyperparameters (number of epochs, etc.) for all training rounds. Finally, GMM selection, has very high performance using as little as 1000 selected images, but the increase is marginal, when more images are selected.

Furthermore, it is worth noting that in the case of selection from Open Images, GMM selection attains almost the same performance as scoring heuristics, but requires 100 times less images. Finally, combining the two selection methods yields better results than using the methods separately for every amount of selected images, except for the 1k case.

\begin{table}
	\setlength\tabcolsep{2.5pt}
	\caption{Performance (\textbf{mIoU}) on Cityscapes Dense classes that receive extra supervision from Open Images.}
	\label{tab:perf-selec-from-open}
	\begin{center}
		\begin{tabular}{c|cccc}
			& \multicolumn{4}{c}{\# of selected images images}\\
			Method of selection & 1k (0.1\%) & 10k (1\%) & 20k (2\%) & 100k (10\%)\\
			\hline
			random & 67.05 & 67.68 & 68.51 & 67.88\\
			heuristics & 65.97 & 67.45 & 68.01 & 68.88\\
			GMM & 68.67 & 68.52 & 68.88 & 69.00\\
			heuristics + GMM & 68.03 & 68.92 & 69.15 & 69.23\\
		\end{tabular}
	\end{center}
\end{table}

\begin{table}
	\setlength\tabcolsep{4.0pt}
	\caption{Detailed per class \textbf{IoU} for the GMM selection method using weakly labeled data from Open Images. Only the 8 out of 19 Cityscapes classes are shown that receive extra supervision.}
	\label{tab:perf-detail}
	\begin{center}
		\begin{tabular}{c|cccccccc}
			\# of images & \rot{car} & \rot{truck} & \rot{bus} & \rot{train} & \rot{motorcycle} & \rot{bicycle} & \rot{person} & \rot{rider}\\
			\hline
			1k & 92.2 & 68.2 & 76.9 & \textbf{71.2} & 50.7 & 67.5 & 71.7 & 51.0\\
			10k & 92.2 & 69.7 & 79.3 & 65.6 & 48.2 & 67.7 & 71.6 & 51.2\\
			20k & \textbf{92.5} & \textbf{73.1} & \textbf{79.9} & 60.9 & \textbf{53.3} & \textbf{67.7} & 71.8 & 52.0\\
			100k & 92.4 & 69.6 & 78.8 & 67.8 & 51.1 & \textbf{68.0} & \textbf{71.9} & \textbf{52.6}
		\end{tabular}
	\end{center}
\end{table}

\subsection{Training with Cityscapes Dense and Open Images}
\label{ssec:perf-open}
Open Images is a completely different dataset than Cityscapes. It contains images from a variety of generic natural scenes, and the street scene images are very limited \cite{kuznetsova2018open}, thus the \textit{domain gap} between them is large, as can be seen also from Figure~\ref{fig:stats-logprobs}. Open Images is labeled with 600 classes, the majority of which are not relevant for automated driving. We experiment with both of our selection methods, since both diversity for street scene classes and \textit{visual similarity} with Cityscapes Dense is needed.

Table~\ref{tab:perf-selec-from-open} shows the detailed mIoU performance on Cityscapes Dense, when the hierarchical model is trained on per-pixel labels of Cityscapes Dense and on various amounts of selected images from per-bounding-box labels of Open Images. In the first row, the mIoU for random selection is shown, which represents a strong baseline. In the second and third row, the proposed techniques of Section~\ref{sec:method} are studied. We observe that selection through GMM has higher gain in small amount of weakly labeled images, while selection with scoring heuristics is better when using more that 20000 weakly labeled images.

Moreover, we investigate the option of combining both selection methods, so we have high \textit{visual similarity} and \textit{object diversity}. The final collection of images is obtained by selecting the same amount from each of the two rankings so each method contributes half of the selected images, after removing duplicate images. Interestingly, the two selection methods have dissimilar rankings, as can be seen in the analysis~\ref{ssec:common-images}.

Finally, we observe that for different number of available images with weak labels, a different selection method is more suitable. Knowing that Cityscapes Dense has 2975 training images, we observe that, if only 1000 weakly labeled images are available, then selecting similarity (GMM) over diversity (heuristics) works better, and the model does not overfit. If weakly labeled images are 100 times more, then selecting \textit{object diversity} gives better results.

In Table~\ref{tab:perf-detail} we present the detailed per class IoU for the GMM selection method. Three classes (car, bicycle, and person) have little gain in performance from the increase in the number of selected images. Four classes (truck, bus, motorcycle, and rider) have significant gain in performance. A potential reason for both groups can be the different point of view that these classes are depicted in the images of Cityscapes and Open Images.

The most interesting case is the train class. We observe that as we include more images until 20k images, the IoU drops dramatically and rises back to a satisfactory level only when using 100k images. This clearly signifies that although the images including trains may appear \textit{visually similar} in whole, the trains between Cityscapes and Open Images have completely different appearance. This remarks the need of investigating \textit{visual similarity} per class instead of per image, but we leave this for future research.

\begin{table}
	\setlength\tabcolsep{3.5pt}
	\caption{Performance (\textbf{mIoU}) on Cityscapes Dense classes that receive extra supervision from Cityscapes Coarse.}
	\label{tab:perf-from-citys}
	\begin{center}
		\begin{tabular}{c|ccc|c}
			& \multicolumn{4}{c}{\# of selected images images}\\
			Method of selection & 1k (5\%) & 5k (25\%) & 10k (50\%) & 20k (100\%)\\
			\hline
			random & 66.68 & 66.82 & 69.38 & \multirow{3}{*}{69.87}\\
			heuristics & 66.61 & 67.58 & 68.77 &\\
			heuristic + GMM & 68.37 & 68.29 & 67.41
		\end{tabular}
	\end{center}
\end{table}

\subsection{Training with Cityscapes Dense and Cityscapes Coarse}
\label{ssec:perf-citys}
Cityscapes Coarse is a subset of Cityscapes, and thus is \textit{visually similar} to Cityscapes Dense, where performance is evaluated on. Thus, through this experiment we can examine the selection method aiming for \textit{object diversity} in isolation, however for completeness we present also results from combining both selection methods. Table~\ref{tab:perf-from-citys} illustrates that our selection methods are useful when using few images from the weakly labeled dataset, but there is no applicability for using more than 10k images. The significant performance drop for the third column with 10k images, is expected and is due to our chosen scheme of scoring heuristics. Specifically, by examining the per class IoU results, we discover that the performance drop is proportional to the scores we assigned during the class scoring heuristics. Moreover, as can be seen from the last row of Table~\ref{tab:perf-from-citys} GMM selection does not add much since \textit{visual similarity} is already attained by using same dataset.

\begin{figure}
	\centering
	\includegraphics[width=1.0\linewidth]{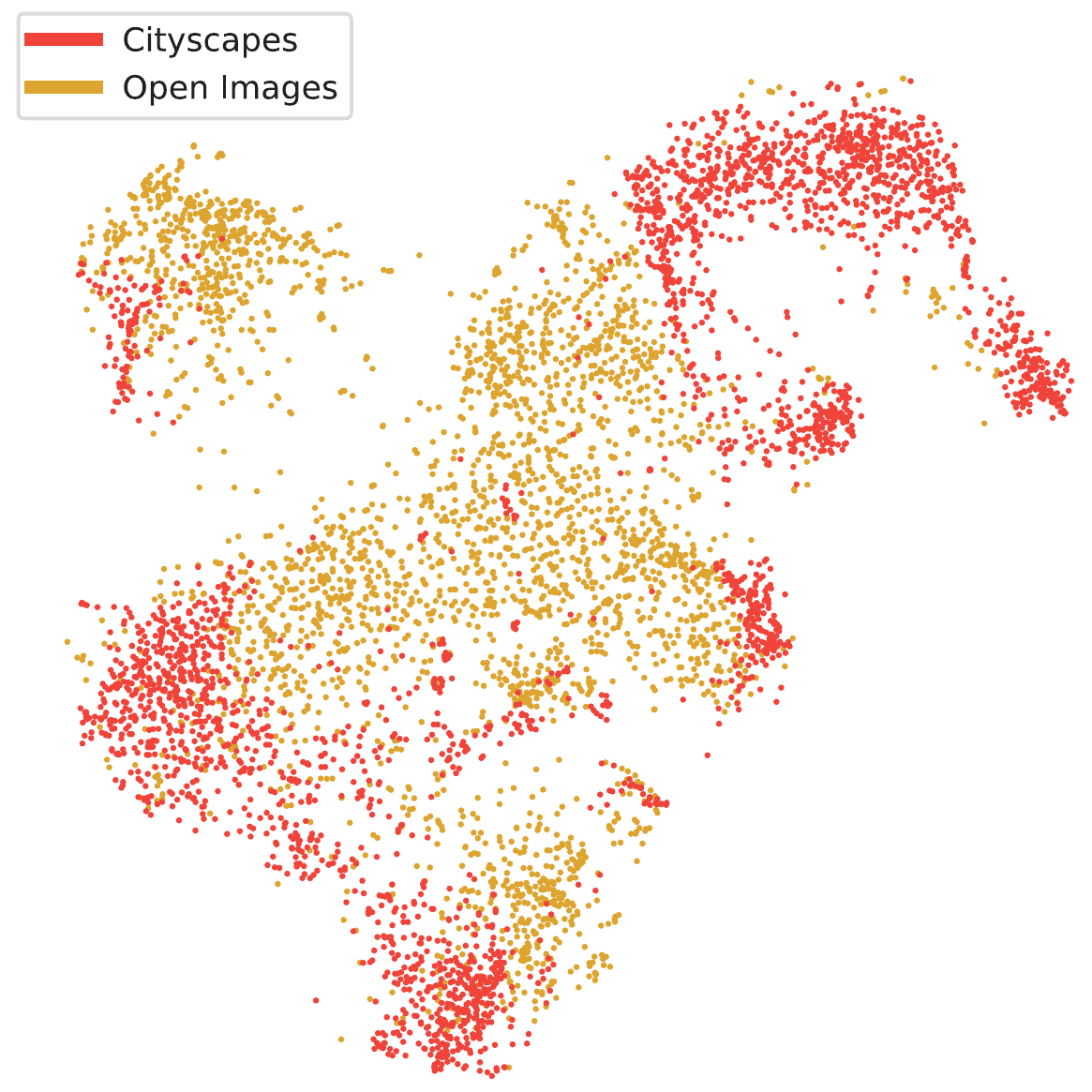}
	\caption{tSNE plot for the image representations for a sample of Cityscapes dense and $1.74$ million Open Images images.}
	\label{fig:tsne-citys-open}
\end{figure}

\subsection{Analysis and ablation experiments}
\label{ssec:analysis}
In this Section we present results towards characterizing the image domains of the datasets we used and perform ablation experiments on the number of components $ K $ of the GMM model.

\subsection*{Dataset characterization and visual similarity}
\label{ssec:dom-model}
Figure \ref{fig:tsne-citys-open} shows the tSNE embeddings~\cite{maaten2008visualizing} of the 256-dimensional image representations $\bar{\Phi}$ for a sample of images from the two datasets. For both datasets, we sampled 3000 representations randomly using our model on the respective training sets. We observe that the distribution of representations have minimal overlap. This separation explains why the GMM can fit the representations from Cityscapes so well and single out representations from Open Images that are dissimilar.

\begin{figure}
    \centering
    \includegraphics[width=1.0\linewidth, trim={0.5cm 12.5cm 0.5cm 4.0cm}, clip]{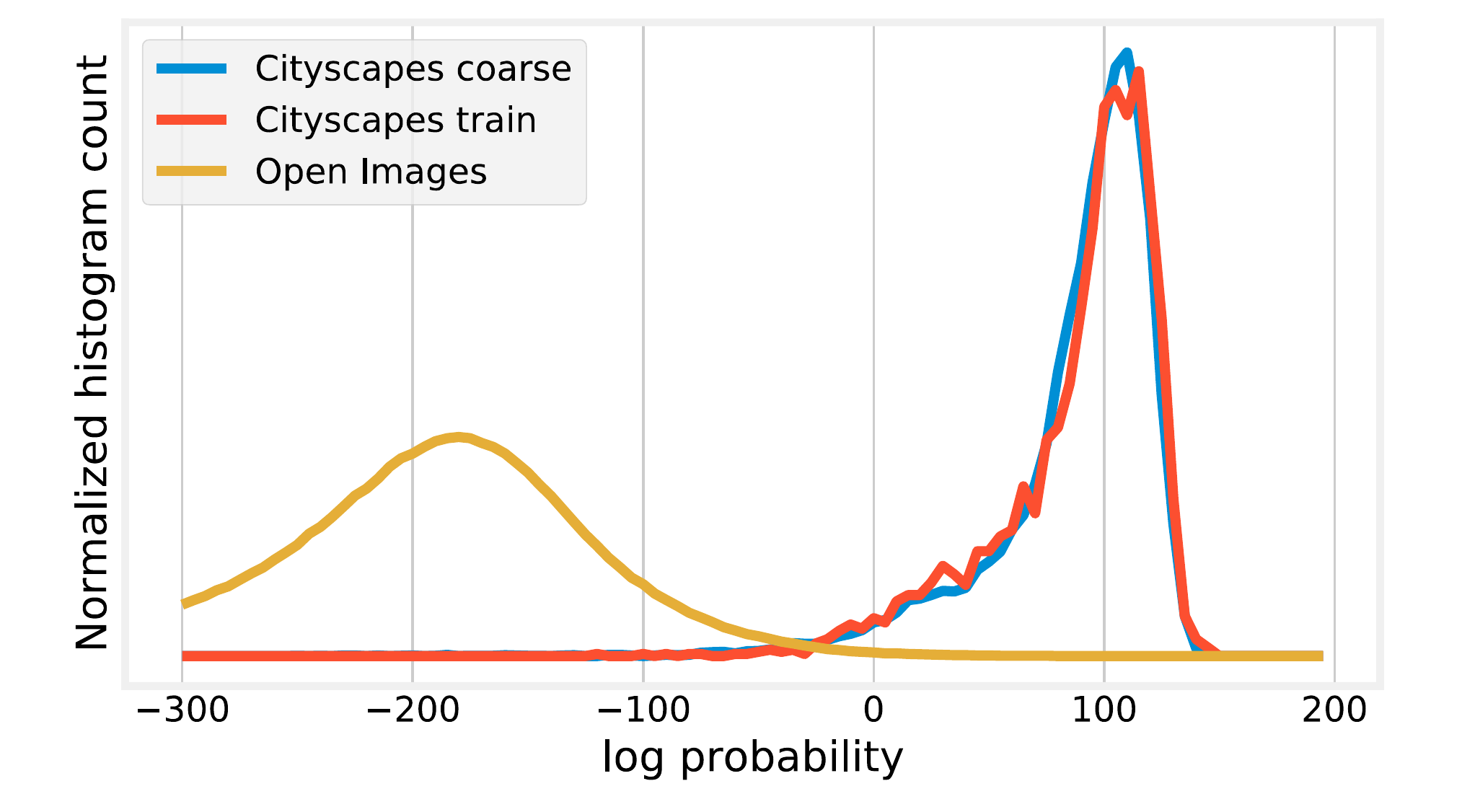}
    \caption{Empirical histogram of the log probabilities for the three datasets. For each image in the training set, we calculate the max log probability and count it to the histogram.}
    \label{fig:stats-logprobs}
\end{figure}

Figure~\ref{fig:stats-logprobs} illustrate statistics of the \textit{visual similarity} measure defined in Section~\ref{sec:method},~\ie max log probability of the GMM, for all images of the three used datasets. From Figure~\ref{fig:stats-logprobs} it can be seen that the histogram of the max log probability for Cityscapes Dense and Coarse image subsets are very similar and confirms their common origin as subsets of Cityscapes. On the contrary, the histogram of Open Images is more spread and has a very small overlap with Cityscapes Dense. The spread of Open Images histogram shows that the scene variety is high, and only a small subset is \textit{visually similar} to Cityscapes. The difference confirms our hypothesis that the images from Open Images follow a different distribution.

\subsection*{Number of GMM components}
In this ablation experiment we investigate the optimal number $ K $ of GMM components for modeling the Open Images domain, having as a indirect metric the performance on Cityscapes validation. As can be seen from Table~\ref{tab:gmm-comp} $ K = 5 $ gives the higher mIoU. The selection of this parameter follows the intuition that representations have a simple and compact structure, as indicated by the tSNE plot of Figure~\ref{fig:tsne-citys-open}, and is guided by the Bayesian Information Criterion (BIC).

\begin{table}
\caption{Ablation on the number of components of GMM for the mIoU performance using the Open Images as the weak dataset.}
\label{tab:gmm-comp}
\begin{center}
\begin{tabular}{c|c|c|c}
K components & 5 & 20 & 50\\
\hline
&&&\\[-1em]
BIC ($\cdot 10^6$) $\downarrow$ & 3.2 & 50.7 & 124.7\\
\hline
&&&\\[-1em]
mIoU $\uparrow$ & \textbf{68.67} & 65.86 & 64.18
\end{tabular}
\end{center}
\end{table}

\subsection*{Common images in rankings}
\label{ssec:common-images}
When both rankings from different selections methods are used (experiment in Table~\ref{tab:perf-selec-from-open}), a conflict of ranking position arises, on which we opted for each ranking to contribute half images to the final selection. Here we compute the agreements of the two ranking approaches. From Table~\ref{tab:common} it can be seen that the two selection methods have different preferences and also that \textit{visual similarity} does not induce \textit{object diversity} and vice verse. Specifically, we were surprised to find out that in 1000 selected images only 0.3\% were selected in the top 500 from both methods.

\begin{table}
\setlength\tabcolsep{3.5pt}
\caption{Common images selected by two techniques.}
\label{tab:common}
\begin{center}
\begin{tabular}{c|cccccc}
\# of selected images & 1k & 10k & 20k & 50k & 100k & 200k\\
\hline
\# of common images & 3 & 117 & 385 & 1492 & 4303 & 10704\\
percentage & 0.3\% & 1.17\% & 1.93\% & 2.98\% & 4.3\% & 5.35\%
\end{tabular}
\end{center}
\end{table}

\section{Discussion and future work}
\label{sec:discussion}
Although we apply our data selection methodology in selecting weak supervision data for multiple dataset semantic segmentation training, it can be used in a variety of problems, such as choosing image for selective per-pixel annotation, balancing training data between datasets, and selection from multiple datasets for semantic segmentation which we will explore in future work.

Another point of discussion concerns using the GMM to capture the representation manifold. We have no grounded reasons to assume that the representations follow local Gaussian distributions. We have attempted to use dimensionality reduction techniques, like PCA and tSNE, but they give no guarantees. To better capture the representation manifold of a particular domain, we might consider more flexible distribution approximators, like normalizing flows\cite{rezende2015variational}, or kernel density estimators\cite{rosenblatt1956remarks}. Despite these considerations, our approach using GMM's have shown improvements, so we expect even better improvements using more flexible density estimators.

A final point for consideration is to select images using local, per-object \textit{visual similarity} instead of image-level similarity. It is clear from experiments of Sections~\ref{ssec:perf-over},~\ref{ssec:perf-open} that although the selected images are similar and depict street scenes in general, the appearance of objects in them can be completely different that what we wish for.

\section{Conclusion}
We presented two data selection methods targeting \textit{visual similarity} and \textit{object diversity} for the problem of semantic segmentation. We tested both methods in the case of training a convolutional network with strong and weak supervision. The selection methods proved particularly useful for selecting images from the weakly labeled datasets, and dramatically decreased the number of required training images, 20 times for Cityscapes and 100 times for Open Images. Moreover, we took steps for characterizing the visual domain of a dataset by modeling the representations of its images with a Gaussian Mixture Model.

\addtolength{\textheight}{-8cm}   

\bibliographystyle{IEEEtran}
\bibliography{root}

\end{document}